\pdfoutput=1

\documentclass[11pt]{article}

\usepackage[]{acl2023}

\usepackage{times}
\usepackage{latexsym}

\usepackage[T1]{fontenc}

\usepackage[utf8]{inputenc}

\usepackage{microtype}

\usepackage{inconsolata}

\usepackage{xcolor}
\usepackage{amssymb}
\usepackage{amsmath}
\usepackage{graphicx}
\usepackage{bbm}
\usepackage{array,arydshln}
\usepackage{booktabs}
\usepackage{multirow}
\makeatletter
\renewcommand{\maketag@@@}[1]{\hbox{\m@th\normalsize\normalfont#1}}%
\makeatother

\title{{MoNET: Tackle State Momentum via Noise-Enhanced Training \\for Dialogue State Tracking}}

\author{Haoning Zhang$^{1,3}$, Junwei Bao$^2$\thanks{~~Corresponding author: baojunwei001@gmail.com}~, Haipeng Sun$^2$,  \\ \bf Youzheng Wu$^2$, Wenye Li$^{4,5}$, Shuguang Cui$^{3,1, 6}$, Xiaodong He$^2$\\
    $^1$FNii, CUHK-Shenzhen ~ $^2$JD AI Research\\
    $^3$SSE, CUHK-Shenzhen ~ $^4$SDS, CUHK-Shenzhen ~ $^5$SRIBD ~ $^6$Pengcheng Lab\\
    {\tt haoningzhang@link.cuhk.edu.cn, \{wyli, shuguangcui\}@cuhk.edu.cn,} \\ {\tt \{baojunwei, sunhaipeng6, wuyouzheng1, xiaodong.he\}@jd.com}
}

\begin{document}
	\maketitle

	\begin{abstract}

    	Dialogue state tracking (DST) aims to convert the dialogue history into dialogue states which consist of slot-value pairs. 
    	As condensed structural information memorizes all dialogue history, the dialogue state in the previous turn is typically adopted as the input for predicting the current state by DST models.
    	However, these models tend to keep the predicted slot values unchanged, which is defined as \textit{state momentum} in this paper. 
    	Specifically, the models struggle to \textit{update} slot values that need to be changed and \textit{correct} wrongly predicted slot values in the previous turn. 
    	To this end, we propose {MoNET} to tackle state \underline{mo}mentum via \underline{n}oise-\underline{e}nhanced \underline{t}raining. 
    	First, the previous state of each turn in the training data is noised via replacing some of its slot values. 
    	Then, the noised previous state is used as the input to learn to predict the current state, improving the model's ability to \textit{update} and \textit{correct} slot values. 
    	Furthermore, a contrastive context matching framework is designed to narrow the representation distance between a state and its corresponding noised variant, which reduces the impact of noised state and makes the model better understand the dialogue history.
    	Experimental results on MultiWOZ datasets show that MoNET outperforms previous DST methods. 
    	Ablations and analysis verify the effectiveness of MoNET in alleviating state momentum issues and improving the anti-noise ability\footnote{Our code is available at \url{https://github.com/JD-AI-Research-NLP/MoNET}}.

	\end{abstract}
	
	    \begin{figure}[t]
		\centering
		\includegraphics[width=3.0in, height=2.3in]{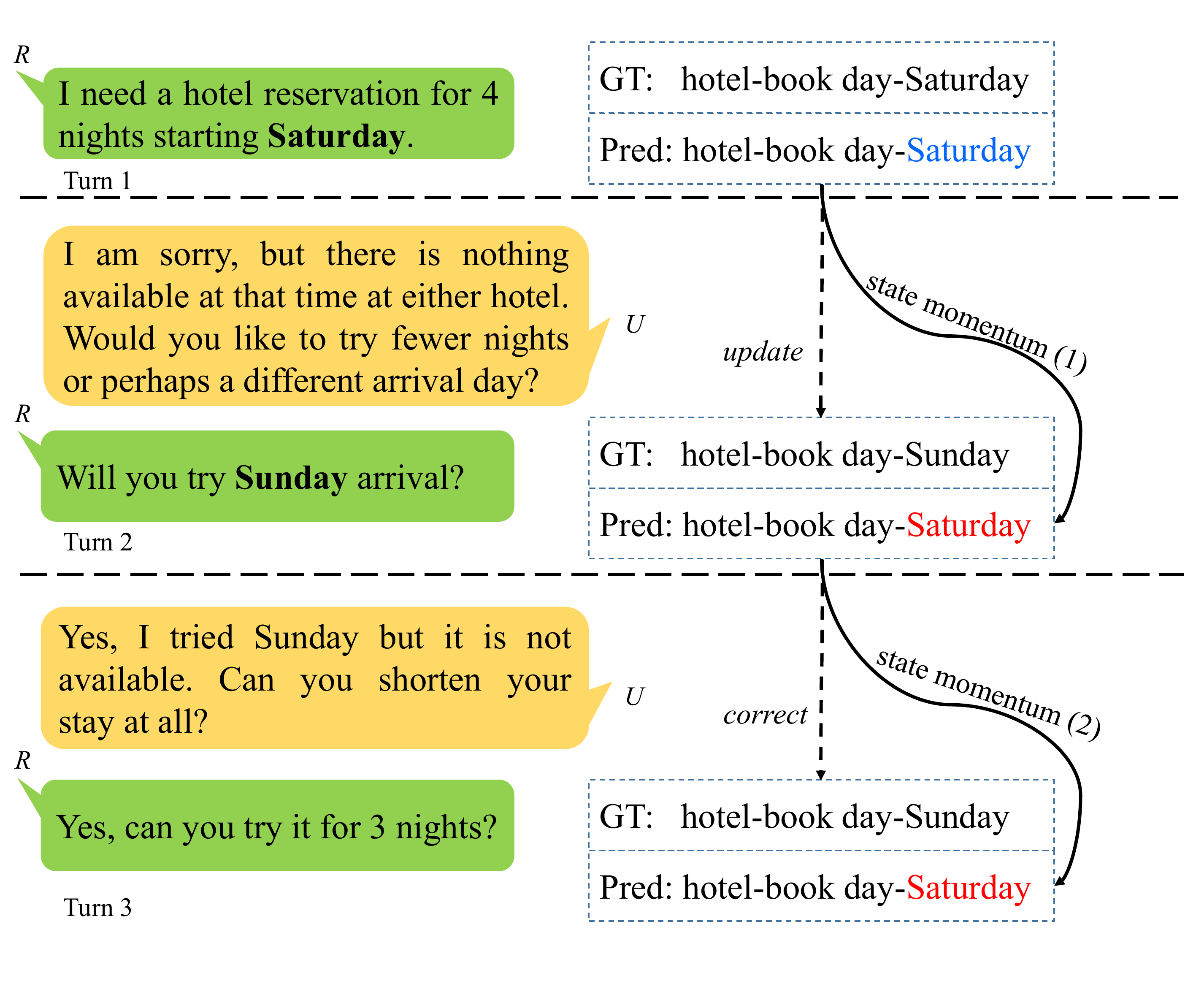}
		\caption{A dialogue example of three turns, containing the system utterance (\textit{U}), the user response (\textit{R}), the ground truth dialogue state (GT), and the prediction of each turn (Pred). The state {``hotel-book day-Saturday''} is predicted in the first turn (marked in \textcolor{blue}{blue}). 
		The dotted arrow represents the ideal predictions, i.e., \textit{update} slot values that need to be changed (Turn 2) and \textit{correct} wrongly predicted slot values in the previous turn (Turn 3).
		The solid arrow represents the predictions (marked in \textcolor{red}{red}) with state momentum issues.
		}
		\label{diagexm}
	\end{figure}
	
	\section{Introduction\label{introduction}}
    Dialogue state tracking (DST) is a core component in modular task-oriented dialogue systems \cite{hosseini2020simple, yang2021ubar, sun-etal-2022-bort,sun-etal-2022-mars}. It extracts users' intents from the dialogue history and converts them into structural dialogue states, i.e., sets of slot-value pairs.
	An accurate dialogue state is crucial for generating correct dialogue action and suitable natural language responses, which are the main tasks of dialogue management and natural language generation components \citep{williams2007partially, thomson2010bayesian, young2010hidden}.
	Earlier DST approaches predict the state directly from the dialogue history (natural language utterances) \citep{mrkvsic2017neural, xu2018end, wu2019transferable, chen2020parallel}.
	Since the dialogue state is condensed structural information memorizing all dialogue history, recent methods incorporate the previously predicted state as the input besides the dialogue history
    \citep{ouyang2020dialogue, kim2020efficient,ye2021slot}. 

    Conventional DST models taking the previous state as the input usually show the characteristic that the previously predicted slot values tend to be kept unchanged when predicting the current state, defined as \textit{state momentum} in this paper.
    The state momentum makes DST models struggle to modify the previous prediction, which affects the performance when the values of some slots need to be \textit{updated} as the user's intent changes, and there exist wrongly predicted slot values that need to be \textit{corrected}. 
    \textcolor{black}{Figure \ref{diagexm} gives an example of a dialogue involving three turns with two types of state momentum issues. The state \textcolor{red}{\textbf{hotel-book day-Saturday}} is predicted in Turn 1 and keeps unchanged in the next two turns, while the user's request is updated into \textbf{Sunday} in Turn 2. Consequently, the predicted state becomes wrong in the following two turns. The dotted arrow represents the ideal prediction cases: the value is \textit{updated} with the ground truth changes and is \textit{corrected} when becoming a wrong input. The solid arrow represents the state momentum issues, where the state is kept unchanged, leading to two consecutive wrong predictions. 
    One possible reason for the state momentum issue is that in the training data, most slot values in the previous turn are the same as those in the current turn, which limits the ability of conventional DST models to modify slot values during inference. To address this limitation, an intuitive idea is to augment training instances with a higher ratio of slots whose previous values differ from those in the current turn. By incorporating such examples, the DST model can learn to deal with more cases where modifying previous predictions is required. Besides, if the DST model can treat wrong and correct dialogue states similarly in representations, then the former will typically help make further predictions. In other words, by treating incorrect dialogue states as valuable information, the DST model can potentially identify and correct erroneous slot values.}

	In this paper, we propose \textbf{MoNET} to tackle the state
    \underline{mo}mentum issue via a \underline{n}oise-\underline{e}nhanced \underline{t}raining strategy. The core idea is to manually add noise into the previous state to simulate scenarios with wrong state input. First, the previous state of each turn in the training data is noised via replacing some of its slot values.
    Specifically, for each active slot (with a non-$none$ value), we replace its value with a certain probability. 
    Then, the noised previous state, concatenated with the dialogue history, is used as the input to learn to predict the current state, improving the model's ability to \textit{update} and \textit{correct} slot values.
    Furthermore, a contrastive context matching framework is designed to narrow the representation distance between a state and its corresponding noised variant, which reduces the impact of the noised state and makes the DST model better understand the dialogue history.
    Such approaches make the model less sensitive to the noise, and enhance its ability to modify the slot values of previous states in current predictions.
	Experiments on the multi-domain dialogue datasets MultiWOZ 2.0, 2.1, and 2.4 show that our MoNET outperforms previous DST models. Ablation studies and analysis further verify the effectiveness of the proposed noised DST training and the contrastive context matching framework in alleviating state momentum and improving the model's anti-noise ability.
	
	
	The contributions are summarized as follows:
	(1) We define the state momentum issue in DST, where models tend to keep the predicted slot values unchanged, namely, struggling to \textit{update} and \textit{correct} them from the previous turn. 
	(2) We propose MoNET to tackle the state momentum issue via noised DST training and the contrastive context matching framework.
	(3) We conduct comprehensive experiments on three datasets, MultiWOZ 2.0, 2.1, and 2.4. The results demonstrate that MoNET outperforms previous DST methods, showcasing its effectiveness in alleviating the state momentum issue.

	\section{Related Work}
	\subsection{Dialogue State Tracking}
	Traditional DST approaches focus on single-domain dialogue state tracking \citep{williams2007partially,thomson2010bayesian,lee2016dialog}.
	Recent researches pay more attention to multi-domain DST using distributed representation learning \citep{wen2017network,mrkvsic2017neural}.
	Previous works implement Seq2seq frameworks to encode the dialogue history, then predict the dialogue state from scratch at every turn \citep{rastogi2017scalable,ren2018towards,lee2019sumbt, wu2019transferable, chen2020parallel}. Utilizing dialogue history is limited for larger turns, since the state of each turn is accumulated from all previous turns, while it's hard to retrieve state information from a long history.
	
	Current works mainly incorporate the previous state as the model input, which is regarded as an explicit fixed-sized memory \citep{ouyang2020dialogue, ye2022assist, wang-etal-2022-luna}.
    \citet{kim2020efficient} propose a state operation sub-task, where the model is trained to first predict the operation of each slot-value pair, such as UPDATE, CARRYOVER, etc., then only the value of a minimal subset of slots will be newly modified
    \citep{zeng2020jointly,zhu2020efficient}. These methods enhance model prediction efficiency and the ability to \textit{update} slot-value pairs.
    \citet{tian2021amendable} deal with the error propagation problem that mistakes are prone to be carried over to the next turn, and design a two-pass generation process, where a temporary state is first predicted then used to predict the final state, enhancing the ability to \textit{correct} wrong predictions.
    In this paper, we use ``state momentum'' to define the issue where the wrong dialogue state is predicted due to that the previous prediction keeps unchanged, either it should be \textit{updated} or \textit{corrected}. 
    To the best of our knowledge, this is the first time to systematically tackle the issue caused by continuous unchanged predictions in the multi-turn DST task.

	\subsection{Contrastive Learning}
	Contrastive learning aims to generate high-quality representations by constructing pairs of similar examples to learning semantic similarity \citep{mnih2012fast,baltescu2015pragmatic,peters2018deep}.
	The goal is to help the model semantically group similar instances together and separate dissimilar instances.
	During training, the neighbors with similar semantic representations (\textbf{positive pairs}) will be gathered, while the non-neighbors (\textbf{negative pairs}) will be pushed apart, enabling the learning of more meaningful representations.
    In the NLP area, semantic representations can be learned through self-supervised methods, such as center word prediction in Word2Vec, next sentence prediction in BERT, sentence permutation in BART, etc \citep{mikolov2013efficient, devlin2019bert, lewis2020bart}.
	Recent approaches build augmented data samples through token shuffling, word deletion, dropout, and other operations \citep{cai2020group,klein2020contrastive,yan-etal-2021-consert,wang2021cline,gao2021simcse, zhang-etal-2022-css}.
	In this paper, we construct augmented samples based on the noised and original dialogue state. Given context inputs with the same dialogue history and different states, the model is trained to gather them into similar objects, aiming to learn better representations, reduce the impact of noise, and better understand the dialogue history.
	
	\section{Methodology}

    \begin{figure*}[ht]
		\centering
			\includegraphics[width=15.5cm]{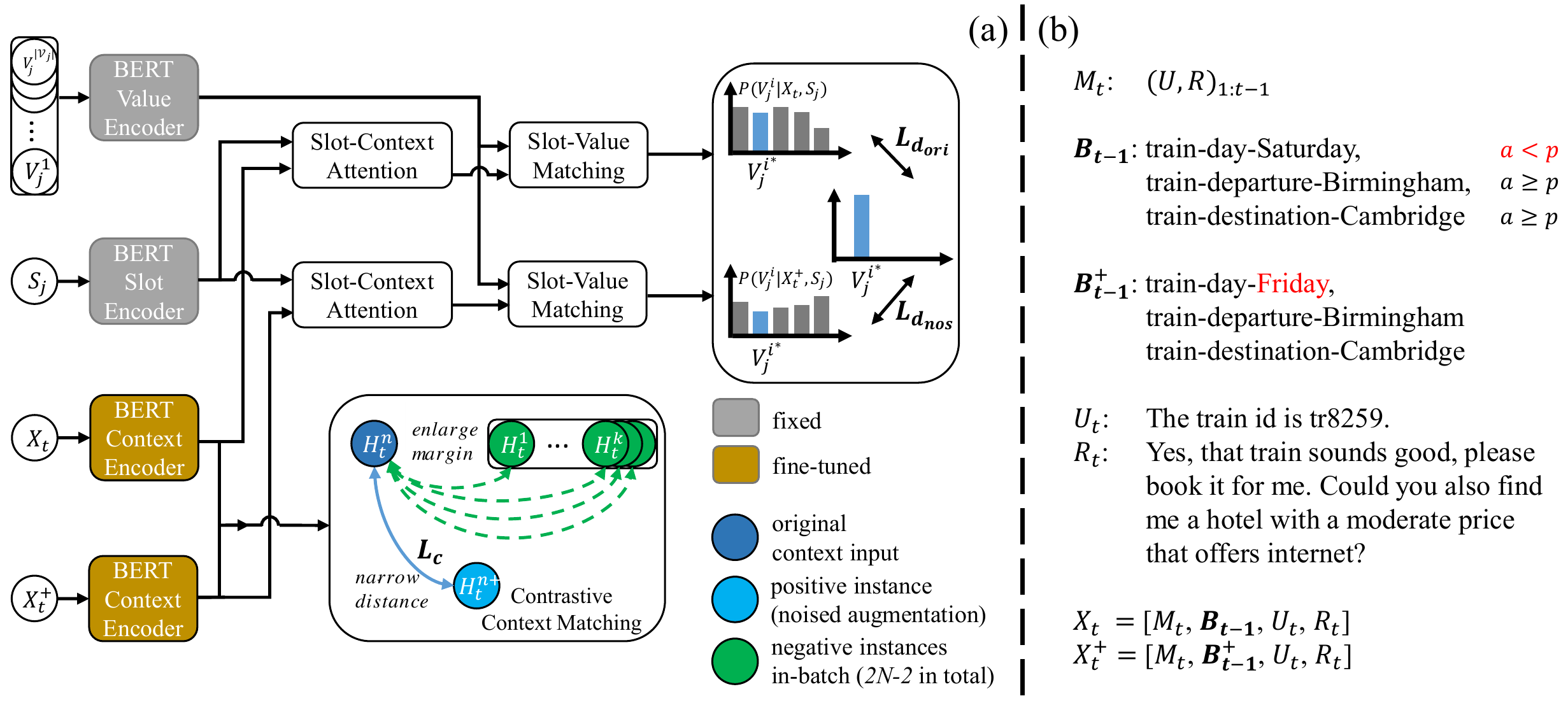}
            \caption{The model description and noised input construction example. The left part (a) shows the architecture of the MoNET model. A context input representation $H^n_t$ in an $N$-size batch is shown in the contrastive context matching framework as an example, where $H^{n+}_t$ is the context representation of its noised variant. 
            The right part (b) gives an example of constructing noised context input. For each active slot in state $B_{t-1}$, given a noise threshold $p$, a random number $a$ is selected. If $a < p$, then its value is replaced with another one randomly selected from the ontology (suppose the pair ``train-day-Saturday'' is replaced with ``train-day-\textcolor{red}{Friday}''); otherwise the value will be kept unchanged (suppose the pairs ``train-departure-Birmingham'' and ``train-destination-Cambridge'' are unchanged).}
			\label{fig:STC}
		\end{figure*}

	\subsection{Problem Formulation}
	In this paper, we focus on building a dialogue state tracking (DST) model which accurately predicts the dialogue state based on the dialogue history and the previous state during multi-turn dialogue interactions.
	A dialogue state consists of domain-slot-value tuples, typically corresponding to the dialogue topic, the user's goal, and the user's intent. 
	Following previous studies, in the rest of this paper, we omit ``domain'' and use ``slot'' to refer to a ``domain-slot'' pair.
	All slot-value pairs are from a pre-defined ontology. 
		
	Formally, let's define $D_t = [U_t, R_t]$ as a pair of system utterance $U_t$ and user response $R_t$ in the $t$-th turn of a multi-turn dialogue, and $B_t$ as the corresponding dialogue state.	
	Each state $B_t$ contains a set of slot-value pairs, i.e., $B_t=\{(S_j,V^{i}_{j}) | j \!\in\! [1\!:\!J]\}$, where $J$ is the total number of slots, and $V^{i}_{j} \in \mathcal{V}_j$ is one of the values in $\mathcal{V}_j$ for the $j$-th slot $S_j$ in the ontology. 
	Given the dialogue history $\{D_1,..., D_t\}$ and previous state $B_{t-1}$, the goal of the DST task is to predict the current dialogue state $B_t$.

	

	\subsection{MoNET}
	As introduced in Section~\ref{introduction}, solving the \textit{state momentum} issue is crucial to the DST task. 
	Therefore, in this paper, we propose MoNET to tackle the state
    \underline{mo}mentum issue via a \underline{n}oise-\underline{e}nhanced \underline{t}raining strategy to enhance the model's ability to \textit{update} and \textit{correct} slot values. 
    The architecture of MoNET is shown in Figure \ref{fig:STC}(a), which consists of context BERT encoders, slot and value BERT encoders, the slot-context attention module, the slot-value matching module, and the contrastive context matching framework. Each of them will be elaborated on in this section.
	
	\subsubsection{Base Architecture}
	We first introduce the base architecture of our MoNET, similar to the backbone model in \citep{ye2022assist}. 
	A model trained only with the base architecture of MoNET is noted as ``Baseline'', and evaluated in Section \ref{5} to compare the difference in performance with the whole MoNET model.
	\label{base_arc}
	\paragraph{Context Encoder.} A BERT encoder encodes the context input, which is the concatenation of the dialogue history and the state in the previous turn:
	\begin{equation}
	\resizebox{0.98\hsize}{!}{
	$
	\begin{aligned}
	&X_t = f(M_t, B_{t-1}, D_t) \\ 
	&= [CLS] \oplus M_t \oplus B_{t-1} \oplus [SEP] \oplus D_t \oplus [SEP],
	\end{aligned}
	$}
	\end{equation}
	where $M_t=D_1 \oplus ,...,\oplus D_{t-1}$ contains previous utterances, $B_{t-1}$ is the state containing the active slots in the previous turn, $[CLS]$ and $[SEP]$ are special tokens of the BERT encoder. Then the representations of the context input are derived:
	\begin{equation}
		H_{t} = BERT(X_t) \in \mathbb{R}^{|X_t| \times d},
	\end{equation}
    where $|X_t|$ is the total number of tokens in $X_t$, and $d$ is the encoded hidden size.
	\paragraph{Slot and Value Encoders.} 
	The BERT encoders with fixed parameters are used to derive the slot and value representations:
	\begin{equation}  
	\begin{aligned}
		&h_{S_j}=BERT_{fixed}(S_j)_{[CLS]}, \\
		&h_{V^{i}_{j}}=BERT_{fixed}(V^{i}_{j})_{[CLS]},
	\end{aligned}
	\end{equation}
	where states $h_{S_j}, h_{V^{i}_j} \in \mathbb{R}^{1 \times d}$ are the $[CLS]$ representations of the slot and value.
	\vskip 0.5em \noindent
	\paragraph{Slot-Context Attention.} 
	For each slot $S_j$, its slot-context-specific feature is extracted by the multi-head attention mechanism \citep{vaswani2017attention}: 
	\begin{equation}
	    r^t_{S_j}=LN(MultiHead(h_{S_j}, H_t, H_t)) \in \mathbb{R}^{1 \times d},
	\end{equation}
	where $LN$ is the normalization layer. 
	\paragraph{Slot-Value Matching.} 
	The probability of predicting the value $V^{i}_{j}$ of the slot $S_j$ is derived by calculating the $L2$-distance between the value representation $h_{V^{i}_{j}}$ and the slot-context representation $r^{t}_{S_j}$, which is denoted as:
	\begin{equation}
	\resizebox{0.95\hsize}{!}{
		$P_{\theta}(V^{i}_{j}|X_t, S_j) = \frac{exp(-||r^{t}_{S_j}-h_{V^{i}_{j}} ||_2)}{\sum_{k \in [1:|\mathcal{V}_j|]}exp(-||r^t_{S_j}-h_{V^{k}_{j}}||_2)}$,}
	\end{equation}
	where $\theta$ are trainable parameters of the model.
	\paragraph{Training and Inference.} 
	During training, the ground dialogue state is used to form the context input $X_t$ (teacher-forcing). For the $t$-th turn, the loss is the sum of the negative log-likelihood among all $J$ slots as follows:
	\begin{equation} 
		L_{d_{ori}} = \sum_{j=1}^J -log(P_{\theta}(V^{i^{*}}_{j}|X_t, S_j)),
	\end{equation} 
	where $V^{i^{*}}_{j}$ is the ground truth value of the slot $S_j$ at turn $t$.
	During inference, the previously predicted state is used to form the context input $X_t$, and the value of the slot $S_j$ is predicted by selecting the one with the smallest distance, corresponding to the largest probability:
	\begin{equation} 
	    {V^{\widehat{i}}_{j}} = \mathop{argmax}_{i \in [1:|\mathcal{V}_j|]} P_{\theta}(V^{i}_{j}|X_t, S_j).
	\end{equation} 

	
    \subsubsection{Noised Data Construction}
    \label{nos_thresh}
        \textcolor{black}{As described previously, an intuitive idea to tackle the state momentum issue is to increase the number of training instances where the slot-value pairs in the previous turn are different from those in the current turn. Based on this point, we attempt to utilize noised data to train the DST model.}
    
        Generally, for each active slot (with a non-$none$ value) in the previous dialogue state, we involve noise by replacing its original value with another value with a probability $p$ (used as the noise threshold), e.g., as the example shown in Figure \ref{fig:STC}(b).
    	Formally, at each training step, given a batch of training instances, a noised context input $X^+_{t}$ is constructed for each instance based on its original context input $X_t=f(M_t, B_{t-1}, D_t)$ as follows:
    	\begin{equation}
    		\begin{aligned}
    		     X^+_{t}=& f(M_t, B^+_{t-1}, D_t), \\
    		  B^+_{t-1}=&\{(S_j,V^{i^+}_{j}) | j \!\in\! [1\!:\!J]\}.
    		\end{aligned}
    		\label{lable_pos_sample}
    	\end{equation}
    	For each active slot $S_j$ in $B_{t-1}=\{(S_j,V^{i}_{j})\}$, a real number $a\in[0,1]$ is sampled to determine whether the original $V^{i}_j$ is replaced with a randomly selected value $V^{k}_{j} \in \mathcal{V}_j \setminus \{V_{j}^{i}\}$ from the ontology or kept unchanged:
        \begin{equation}
        \begin{aligned}
                V^{i^+}_{j}= \left \{
                    \begin{array}{ll}
                        V^{k}_{j}, &if \quad a < p \\
                        V^{i}_{j}, &if \quad a \geq p.
                    \end{array}
                        \right.
            \end{aligned}
        \end{equation}

    
    \subsubsection{Noised State Tracking}	
	\label{noised_DST}
    Similar to $X_{t}$, the noised context instance $X_{t}^+$ is also used as the model input to predict the state $B_{t}$ as the training target, aiming to improve the model's ability to dynamically modify the previous slot values in current predictions.
	Specifically, the representation $H_t^+$ of $X^+_t$ is first derived by the BERT context encoder mentioned in Section \ref{base_arc}:
	 \begin{equation}
	    H_t^+ = BERT(X^+_{t})  \in \mathbb{R}^{|X^+_{t}| \times d}.
	 \end{equation}
	Then, similar to the previous process, for each slot $S_j$, $X^+_t$ is used to predict its value based on the distribution $P_{\theta}(V^{i}_{j}|X^+_t, S_j)$. 
	Eventually, the loss for the noised state tracking can be denoted as:
	\begin{equation} 
		L_{d_{nos}} = \sum_{j=1}^J -log(P_{\theta}(V^{i^*}_{j}|X^+_t, S_j)).
	\end{equation} 
		
	\subsubsection{Contrastive Context Matching}
		
    Inspired by contrastive learning approaches which group similar samples closer and diverse samples far from each other, a contrastive context matching framework is designed to narrow the representation distance between $X_t$ and its noised variant $X_t^+$, aiming to reduce the impact of the noised state $B_{t-1}^+$ and help the model better understand the dialogue history.
	Specifically, in a batch of $N$ instances with the original context input $\mathbf{X}_t = \{X_t^n\}_{n=1}^{N}$, we construct $N$ corresponding noised instances with the context input $\mathbf{X}_t^+ = \{X_t^{n+}\}_{n=1}^{N}$.
    To clearly describe the context inputs, in this section, we temporarily involve $n$ into $X_t$ \& $H_t$ as $X_t^n$ \& $H_t^n$ to indicate the in-batch index.
 
	For each context input $X_t^n$, its noised sample $X_t^{n+}$ is regarded as its positive pair, and the rest $(2N\!-\!2)$ instances in the same batch with different dialogue histories are considered negative pairs.
	Then the model is trained to narrow the distance of the positive pair and enlarge the distance of negative pairs in the representation space with the following training objective:
	\begin{equation}
	    \resizebox{0.98\hsize}{!}{
	    $L_c = -log( \frac{exp(sim(H^{n[cls]}_{t}, H_t^{n+[cls]})/\tau)}{\sum_{k=1}^{2N}\mathbbm{1}_{[k\neq n]}exp(sim(H_t^{n[cls]}, H_t^{k[cls]})/\tau)})$,}
	\end{equation}
	where $H_t^{n[cls]}$ and $H_t^{n+[cls]}$ are the $[CLS]$ representations of $H_t^n$ and $H_t^{n+}$, $\tau$ is the temperature parameter, and $sim(\cdot)$ indicates the cosine similarity function \citep{chen2020simple}.
    \subsubsection{Optimization}
    The total training loss for each instance is the sum of losses from the slot-value matching for DST and the contrastive context matching for representation learning, where the former is the average of the losses using the original or the noised context input mentioned in Section
	\ref{base_arc} and \ref{noised_DST}:
	\begin{equation}
	    L_{tot} = (L_{d_{ori}} + L_{d_{nos}})/2 + L_c.
	\end{equation}
	
	\begin{table*}[!t]
    \centering
    \small
    \resizebox{\textwidth}{!}{
        \scalebox{0.2}{\begin{tabular}{@{}llllllll@{}}
        \toprule
        \multirow{2}{*}{Baseline} & \multirow{2}{*}{Pre-trained Model} & \multicolumn{2}{c}{MultiWOZ 2.0} & \multicolumn{2}{c}{MultiWOZ 2.1} & \multicolumn{2}{c}{MultiWOZ 2.4} \\ \cmidrule(l){3-4} \cmidrule(l){5-6} \cmidrule(l){7-8} 
        &  & Joint & Slot & Joint & Slot & Joint & Slot \\ \midrule
        TRADE \citep{wu2019transferable}  & - & 48.62 & 96.92 & 45.60 & 96.55 & 55.05 &  97.62 \\
        SUMBT \citep{lee2019sumbt}  & BERT-base & 42.40 & - & 49.01 & 96.76 & 61.86 & 97.90            \\
        PIN  \citep{chen2020parallel}   & - & 52.44 & 97.28      & 48.40 & 97.02      & 58.92 & 98.02        \\
        SOM-DST \citep{kim2020efficient}  & BERT-base & 51.72 & - & 53.01 & - & 66.78 & 98.38 \\
        CSFN-DST \citep{zhu2020efficient} & BERT-base & 52.23 & - & 53.19 & - & - & - \\
        DST-Picklist \citep{zhang2020find}  & BERT-base &  54.39 & - &  53.30 & 97.40 & - & -            \\
        SAVN \citep{wang2020slot} & BERT-base & 54.52 & 97.42 & 54.86 & 97.55 & 60.55 & 98.05 \\		
        SST \citep{chen2020schema} & BERT-base & 51.17 & - & 55.23 & - & - & - \\
        SimpleTOD  \citep{hosseini2020simple}  & DistilGPT2 & - & - & 55.26 & - & - &-    \\
        Seq2SeqDU \citep{feng2021sequence}  & BERT-base & - & - & 56.10 & - & - & -  \\ 
        STAR \citep{ye2021slot}      & BERT-base & 54.53 & - & 56.36 & 97.59 & 73.62 & 98.85   \\
        SDP-DST \citep{lee2021dialogue} & T5-base & - & - & 56.66 & - & - & - \\
        DS-Graph \citep{lin2021knowledge} & GPT2 & 54.86 & 97.47 & - & - & - & - \\
        DSGFNet \citep{feng2022dynamic} & BERT-base & - & - & 56.70 & - & - & - \\ 
        PPTOD \citep{su-etal-2022-multi} & T5-large & 53.89 & - & 57.45 & - & - & - \\ \midrule
        Baseline   & BERT-base &  54.38 & 97.47  & 55.82  & 97.51 & 73.81 & 98.82  \\     
    	MoNET   & BERT-base & \textbf{55.48} ($\uparrow$ 1.10)  &  97.55 & \textbf{57.71}  ($\uparrow$ 1.89) & 97.71      & \textbf{76.02}  ($\uparrow$ 2.21) & 98.99       \\ \midrule
        Use Modified Label  \\ \midrule
        TripPy \citep{heck2020trippy} & BERT-base & - & - & 55.29 & - & - & -      \\
        TripPy + SCoRe \citep{yu2020score} & BERT-base & - & - & 60.48 & - & - & - \\
        TripPy + CoCoAug \citep{li2020coco} & BERT-base & - & - & 60.53 & - & - & - \\
        TripPy + SaCLog \citep{dai2021preview}& BERT-base & - & - & 60.61 & - & - & - \\
        \bottomrule[1.3pt]
        \end{tabular}}
    }
    \caption{Joint and slot goal accuracy of our MoNET and several previous methods on three MultiWOZ test sets.} 
	\label{jgamain}
    \end{table*}

	\section{Experiment Setting}
	\subsection{Datasets}
	We choose MultiWOZ, 2.0, 2.1, and 2.4 versions as our datasets.
	MultiWOZ 2.0 \citep{budzianowski2018multiwoz} is a standard human-human conversational dialogue corpus with seven domains.
	MultiWOZ 2.1 \citep{eric2020multiwoz} has the same dialogues as the 2.0 version, where some incorrect state labels are re-annotated.
	Both of them are widely used in previous DST approaches. MultiWOZ 2.4 \citep{ye2021multiwoz} is the latest refined version correcting all the incorrect state labels in validation and test sets. All three datasets contain the same number of dialogues, which are  8438/1000/1000 in train/validation/test sets.
    For the three datasets, we follow the previous work \citep{wu2019transferable} to use five domains (attraction, hotel, restaurant, taxi, train) with 30 domain-slot pairs in experiments, since the dialogues in the remaining domains are not in the validation and test sets. 
	
	\subsection{Evaluation Metrics}
	We use joint and slot goal accuracy as the evaluation metrics. Joint goal accuracy is the ratio of dialogue turns where the values of all slots are correctly predicted. Slot goal accuracy is the ratio of domain-slot pairs whose values are correctly predicted. Both of them include correctly predicting those inactive slots with the value $none$.

    \subsection{Existing Methods}
	We compare the performance of our MoNET with several existing methods, i.e., TRADE, SUMBT, PIN, SOM-DST, CSFN-DST, DST-Picklist, SAVN, SST, SimpleTOD, TripPy, STAR, SDP-DST, DS-Graph, DSGFNet, PPTOD shown in Table~\ref{jgamain}, and our base architecture mentioned in Section \ref{base_arc}, denoted as Baseline.


	\subsection{Training Details}
	The BERT-base-uncased model is used as the context, slot and value encoders, with 12 attention layers and a hidden size of 768.
	During training, only the parameters of the context BERT encoder are updated, while the parameters of the slot and value BERT encoders are not.
	The batch size is set to 8. The AdamW optimizer is applied to optimize the model with the learning rate 4e-5 and 1e-4 for the context encoders and the remaining modules, respectively \citep{loshchilov2018decoupled}.
    The temperature parameter $\tau$ is set to 0.1. 
    The noise threshold $p$ defined in Section~\ref{nos_thresh} is set to 0.3,
    and its impact on model performance is discussed in Section \ref{5}. All models are trained on a P40 GPU device for 6-8 hours.

	\section{Results and Analysis}
	\label{5}
	\subsection{Main Results}
	Table \ref{jgamain} shows performances of MoNET and baselines on MultiWOZ 2.0, 2.1 and 2.4. \textcolor{black}{Among them, TripPy and its modified versions employ a ground truth label map of synonyms replacement as extra supervision, which increases their accuracy scores and differs from other methods of testing with common labels.} As can be observed, MoNET achieves the joint goal accuracy scores of 55.48\%, 57.71\%, 76.02\% in three datasets, which are impressive results compared with previous methods, and has improvements of 1.10\%, 1.89\%, and 2.21\% on the Baseline model, indicating that our proposed noise-enhanced training helps the model make better predictions.

   \textcolor{black}{Besides the general joint and slot goal accuracy, we also calculate the slot-level proportion of state momentum errors over all wrong predictions. We train the Baseline model and make predictions on the MultiWOZ 2.4 test set. For each dialogue, starting from the second turn, we add up each wrong predicted slot-value pair which also exists in the previous turn. Finally, there are 844 such wrong slot-value pairs, and the number of all the wrong predicted pairs is 2603, hence the proportion is (844/2603)*100\%=32.4\%, and our MoNET model modifies 47.0\% of them (397 in 844 are correctly predicted). Moreover, in MultiWOZ 2.4 training set annotations, for each dialogue turn (also except the first turn of each dialogue), around 78.1\% slot-value pairs exist in the previous turn, since the slot-value pairs will be accumulated as the dialogue progresses. The results further indicate the issue caused by those unchanged slot-value pairs during multi-turn interactions, and the effectiveness of our method in enhancing the model's ability to modify previous predictions.}



	\begin{table}[!t]
		\centering
		\small
		\scalebox{0.95}{
		\begin{tabular}{@{}lccc@{}}
			\toprule[1.3pt]
			Model            &NoisedCM  &NoisedST  & Accuracy \\ \midrule
			Baseline w/o state   & $\times$ & $\times$  & 64.94 \\
			Baseline & $\times$ & $\times$ & 73.81 \\ \midrule
			MoNET-ST & $\times$& \checkmark  & 75.54 \\ 
		    MoNET-CM &\checkmark &$\times$ & 75.76 \\ \midrule
			MoNET & \checkmark & \checkmark  & 76.02 \\ \bottomrule[1.3pt]
		\end{tabular}}
		\caption{Joint goal accuracy on the MultiWOZ 2.4 test set of MoNET and four ablated modifications.}
		\label{bttest}
	\end{table}

    \subsection{Ablation Study}
	To explore the individual contribution of each part of our model, we compare the whole MoNET with several ablated versions. First, we remove the previous dialogue state from the context input of the Baseline model,
	where the modified context input is $X_t = [CLS] \oplus M_t \oplus [SEP] \oplus D_t \oplus [SEP]$, denoted as Baseline w/o state;
	besides, the two noise-enhanced methods are removed from the MoNET respectively, denoted as MoNET-CM (context matching only) and MoNET-ST (noised state tracking only). 
 
	Table \ref{bttest} shows the joint goal accuracy performances of the full MoNET model and its four modifications on the MultiWOZ 2.4 test set. As can be observed,
	Baseline w/o state gets the lowest accuracy, demonstrating that explicitly using the previous dialogue state as part of the model input is beneficial to make predictions, even though there may exist wrong slot-value pairs.
	Besides, both MoNET-CM and MoNET-ST outperform the Baseline model, demonstrating functionalities of the noised state tracking in modifying slot-value pairs in further turns, and the context matching framework in learning improved semantic representations. Moreover, MoNET derives the best performance, demonstrating the effectiveness of integrating the two parts into a unified noised-enhance training strategy.
	
	\begin{figure}[!t]
		\centering
		\includegraphics[width=3.0in]{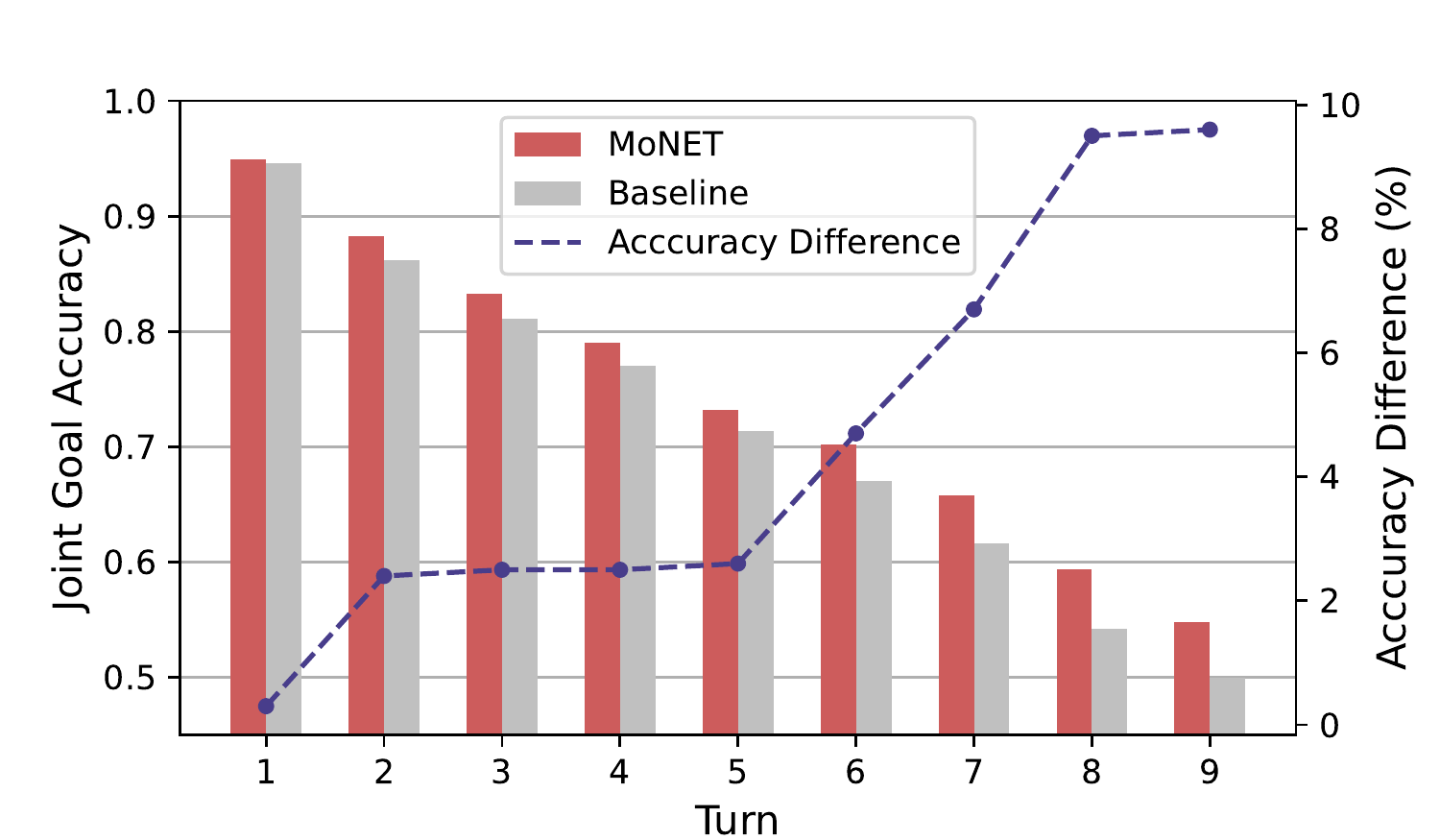}
		\caption{Turn-level joint goal accuracy and accuracy difference between MoNET and Baseline on the MultiWOZ 2.4 test set.}
		\label{turnacc}
	\end{figure}


	\subsection{Turn-Level Evaluation}
	Figure \ref{turnacc} shows the turn-level joint goal accuracy of MoNET and Baseline models, as well as the percentage difference in accuracy (the difference between the two models' accuracy divided by the accuracy of Baseline) on the MultiWOZ 2.4 test set. Generally, the state momentum issue becomes more apparent in dialogues with larger turns, since they always contain more active slot-value pairs, and any one of the wrong pairs kept unchanged will affect the further prediction accuracy. With the increase of turns, the accuracy of Baseline harshly degrades, while MoNET gets a relatively smaller decline, resulting in a gradually increasing and evident percentage difference in accuracy. This demonstrates the superiority of MoNET in alleviating the accuracy decrease caused by the state momentum issue, especially in those dialogues with larger than 6-7 turns. 
	
	\begin{figure}[!t]
		\centering
	 	\includegraphics[width=3.0in]{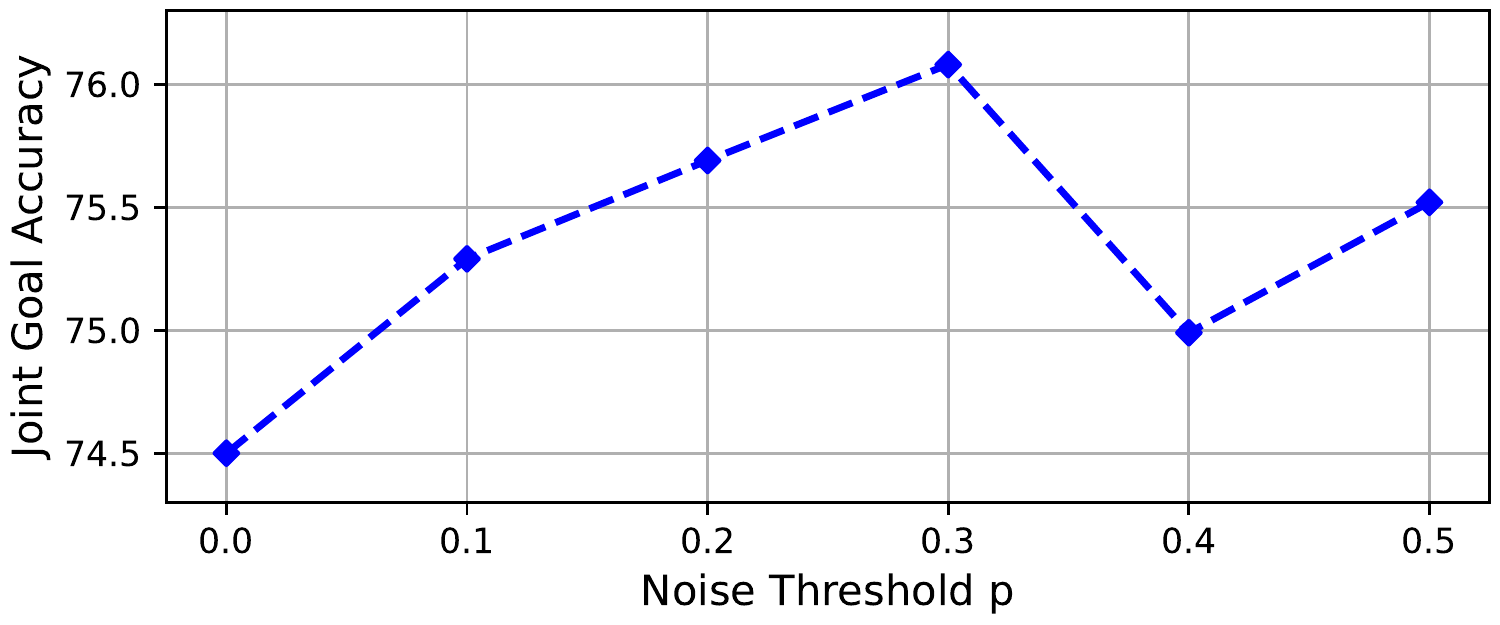}
		\caption{Performance on the MultiWOZ 2.4 validation set w.r.t the noise threshold of adding noise.}
		\label{nosexp}
	\end{figure}
	
	\begin{figure}[!t]
		\centering
	    \includegraphics[width=3.0in]{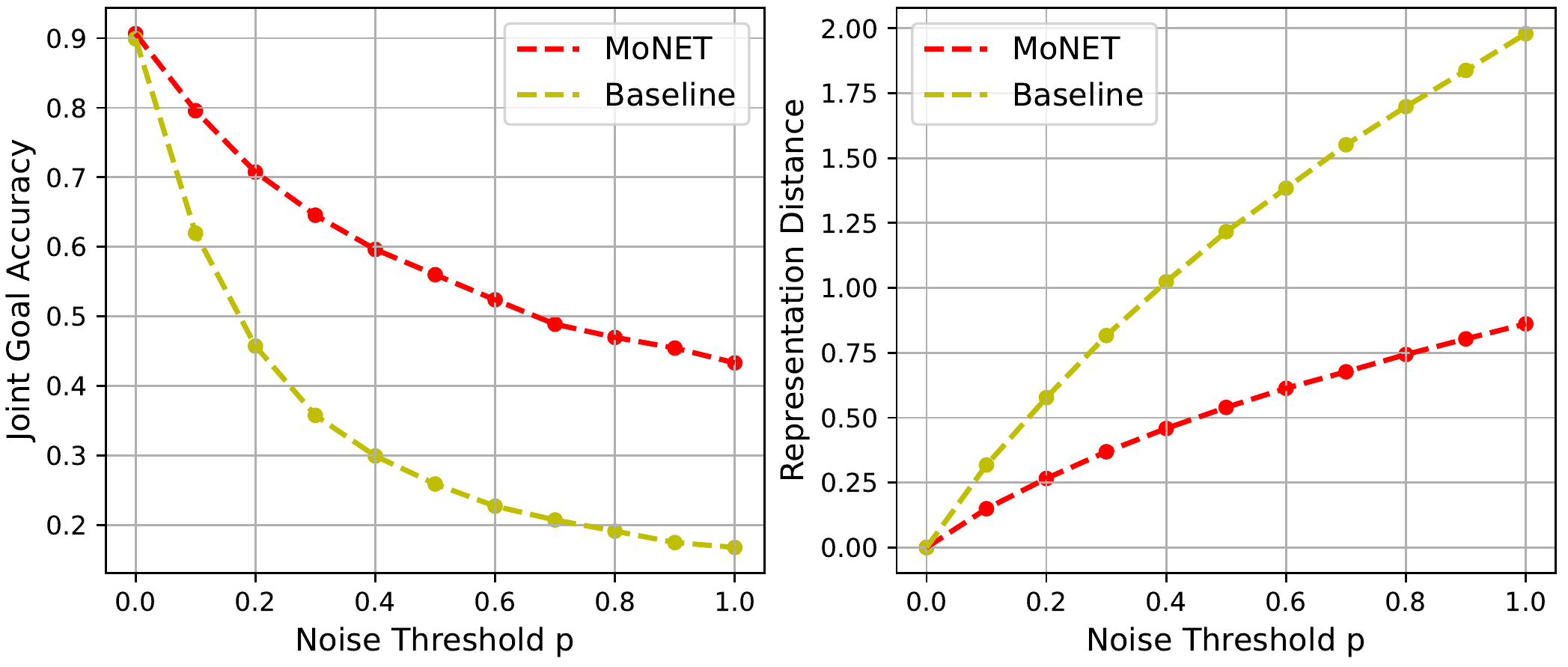}
		\caption{Performances on the MultiWOZ 2.4 test set w.r.t. the noise threshold (corresponding to the noised slot-value pair ratio in the dialogue state input). The left one is the joint goal accuracy, and the right one is the active slot-context features similarity.}
		\label{accsimdeg}
	\end{figure}
	
	\subsection{Noise Threshold Selection for Training}
	To explore the impact of different probabilities of adding noise into the context input for training, we vary the noise threshold $p$ from 0 to 0.5 to train our MoNET. The results on the MultiWOZ 2.4 validation set are shown in Figure \ref{nosexp}, where MoNET achieves the best performance when the noise threshold $p$ is set to 0.3.
	Intuitively, a small $p$ makes the noised context input contain fewer noised slot-value pairs (hard to learn meaningful semantics from the noised data); 
	conversely, a large $p$ makes the noised context input far from the original context input in the representation space (hard to group them closer).
	Both two cases make the model hard to learn effective features from the noised context input, leading to lower prediction accuracy. Hence, the empirical probability of adding noise is important to derive the best performance of the DST model.
	
    \begin{figure*}[!t]
		\centering
		\includegraphics[width=16cm]{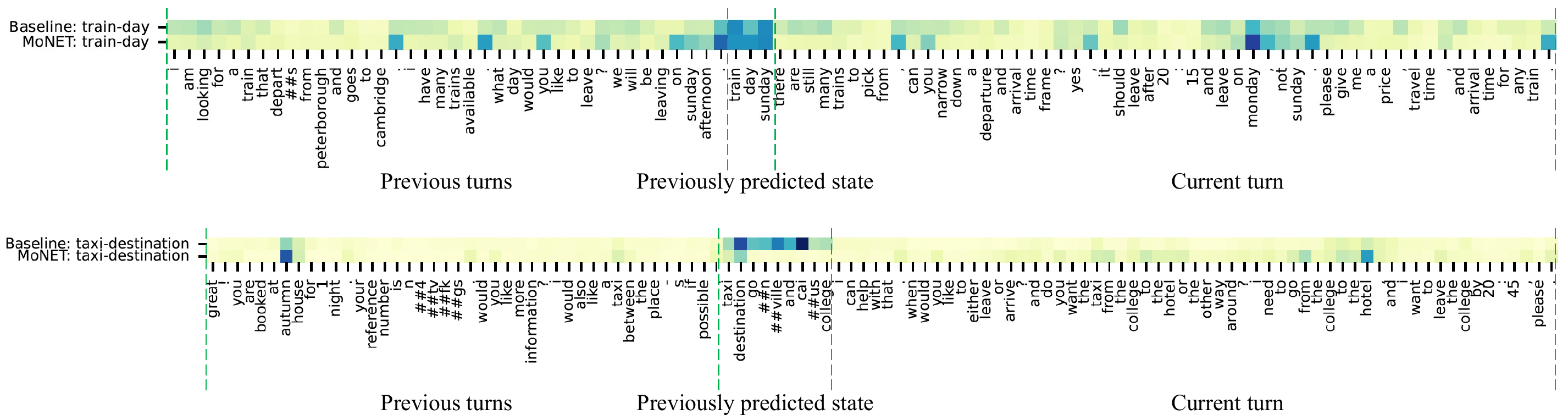}
		\caption{Attention visualizations of the two dialogue examples mentioned in Table \ref{cases}.}
		\label{nosheat}
	\end{figure*}
	
		\begin{table}[!t]
		\centering
		\scriptsize
		\begin{tabular}{@{}l@{}}
			\toprule[1.3pt]
			\begin{tabular}[c]{@{}l@{}}Sys: I have many trains available. What day would you like to leave? \\ Usr: We will be leaving on Sunday afternoon. \\
			GT: train-day-Sunday \\ 
			Baseline: train-day-\textcolor{blue}{Sunday} \qquad 
			MoNET: train-day-\textcolor{blue}{Sunday} \\
			\end{tabular} \\ \midrule
			\begin{tabular}[c]{@{}l@{}}Sys: There are still many trains to pick from, can you narrow down a \\ \qquad departure and arrival time frame? \\ Usr: Yes, it should leave after 20:15 and leave on Monday, not Sunday. \\ \qquad Please give me a price, travel time, and arrival time for any train.\\ 
			GT: train-day-Monday \\
			Baseline: train-day-\textcolor{red}{Sunday} \qquad 
			MoNET: train-day-\textcolor{blue}{Monday} (\textit{update}) \\ 
			 \end{tabular} \\ \midrule[1.1pt] \midrule[1.1pt]
			\begin{tabular}[c]{@{}l@{}}Sys: Great! You are booked at Autumn House for 1 night. Your reference \\ \qquad number is n4tvfkgs. Would you like more information?\\
			Usr: I would also like a taxi between the places if possible. \\
            GT: taxi-destination-Autumn House \\
			Baseline: 
			taxi-destination-\textcolor{red}{Gonville and Caius College} \\
			MoNET: 
			taxi-destination-\textcolor{red}{Gonville and Caius College} \\
			\end{tabular} \\ \midrule 
			\begin{tabular}[c]{@{}l@{}}Sys: I can help with that. When would you like to either leave or arrive?\\ \qquad And do you want the taxi from the college to the hotel or the other \\ \qquad way around? \\
			Usr: I need to go from the college to the hotel, and I want to leave the \\ \qquad college by 20:45, please. \\
			GT: 
			taxi-destination-Autumn House \\
			Baseline: taxi-destination-\textcolor{red}{Gonville and Caius College} \\
            MoNET: taxi-destination-\textcolor{blue}{Autumn House} (\textit{correct})
            
             \end{tabular} \\ 
			\bottomrule[1.3pt]
		\end{tabular}
		\caption{Predictions of two dialogue examples on MultiWOZ 2.4 separated by the double solid line, corresponding to two state momentum cases. Wrong and correct predicted values are marked in \textcolor{red}{red} and \textcolor{blue}{blue}.}
		\label{cases}
	\end{table}

    \subsection{Anti-noise Probing with Noise Testing}
	In this section, we conduct \textit{noise testing} to explore the impact of anti-noise ability on DST models.
	We first evaluate DST performances of MoNET and Baseline by introducing different ratios of noise (with $p$ from 0 to 1) into the oracle previous dialogue state as the model input.
    Figure \ref{accsimdeg} shows the performances of MoNET and Baseline on MultiWOZ 2.4. Both of them get high accuracy when the noise ratio is 0, as we use the oracle previous dialogue state as the model input; 
    with the increase of the noise ratio, the joint goal accuracy of Baseline gets a sharp decline, while MoNET degrades much more smoothly. 
    Furthermore, for each dialogue turn, we also show the $L2$-distance between the original and noised context representations, i.e., the mean pooling of all token representations $H_t$ and $H_t^+$. 
    As can be observed, along with the increase of noise ratio, the distance between the two representations of MoNET is much lower than that of Baseline. These results indicate that MoNET achieves a higher anti-noise ability by generating relatively similar representations for the original and noised contexts, which helps the DST model maintain an acceptable performance even with a high ratio of noise in its input.

\subsection{Case Study and Attention Visualization}
	\label{casestudy}
	\textcolor{black}{Table \ref{cases} gives two prediction examples using MoNET and Baseline on the MultiWOZ 2.4 test set, corresponding to the two types of state momentum cases. In the first one, they correctly predict the slot-value pair ``train-day-\textcolor{blue}{Sunday}'', while only MoNET updates it in the next turn along with the ground truth changing into ``train-day-\textcolor{blue}{Monday}''. In the second one, they make a wrong prediction ``taxi-destination-\textcolor{red}{Gonville and Caius College}''. While Baseline keeps it unchanged in the next turn, MoNET corrects it, resulting in a joint goal accuracy of 100\% for the second turn.
	Besides, we further explore these two examples by calculating and visualizing the overall attention scores, which are shown in Figure \ref{nosheat}. 
    For each slot, its overall attention score over each token is the weighted sum of the self-attended scores by all tokens in $X_t$. The weights come from the slot-context attention, and the self-attended scores are the average of attention scores over multiple layers in BERT. As can be observed, Baseline pays more attention to the values in the previously predicted state, and fails to solve the state momentum issues; MoNET pays relatively higher attention to the correct tokens (``monday'' in the first case and ``autumn house'' in the second case), and consequently, successfully \textit{updates} \textcolor{blue}{Sunday} into \textcolor{blue}{Monday} and \textit{corrects} \textcolor{red}{Gonville and Caius College} into \textcolor{blue}{Autumn House}. These examples and attention visualizations indicate the effectiveness of our MoNET in alleviating the two types of state momentum issues.}

\begin{table}[!t]
\center
\small
    \begin{tabular}{@{}ll@{}}
    \toprule
    Model & Accuracy \\ \midrule
    MinTL \cite{lin2020mintl}  & 52.07 \\
    MTTOD \cite{lee2021improving} & 53.56 \\
    PPTOD  \cite{su-etal-2022-multi} & 53.37 \\ \midrule
    T5-base & 53.26 \\
    MoNET (T5-base) & 54.67 ($\uparrow$ 1.41) \\ \bottomrule
    \end{tabular}
    \caption{Joint goal accuracy on MultiWOZ 2.0 test set of baselines using the same T5-base pre-trained model.}
    \label{generation}
\end{table}

    \subsection{Extension on Generation-based Models}
    \label{T5gen}
        \textcolor{black}{In addition to the original classification-based MoNET model, we also evaluate our approach using a simple generation framework using T5-base as the backbone pre-trained model \cite{raffel2020exploring}. The ontology is built from the database and training set annotations, which is only used for noise value construction. The model framework is similar to the BERT-based MoNET in Figure \ref{fig:STC}(a), where the BERT encoders and slot-value matching modules are replaced with T5 encoders and decoders. The T5 encoders encode the dialogue context inputs, slots, and values. After deriving the slot-context attentive representations, the T5 decoders generate each slot-value pair. Table \ref{generation} shows the joint goal accuracy performance of the T5-based MoNET on the MultiWOZ 2.0 test set, compared with other end-to-end/generation-based models using the same T5-base pre-trainied model. As can be observed, our modified MoNET outperforms the T5-base backbone and others with the same T5-base model, indicating its effectiveness and adaptability for the implementation of generation-based methods.}

    \section{Conclusion}
    \textcolor{black}{In this paper, we define and systematically analyze the state momentum issues in the DST task, and propose MoNET, a training strategy equipped with noised DST training and the contrastive context matching framework. Extensive experiments on MultiWOZ 2.0, 2.1, and 2.4 datasets verify its effectiveness compared with existing DST methods. Supplementary studies and analysis demonstrate that MoNET has a stronger anti-noise ability which helps alleviate the state momentum issues.}
 

    \section*{Limitations}	
	\textcolor{black}{Our proposed MoNET is a classification-based method requiring the pre-defined ontology containing all slot-value pairs. Moreover, during prediction, for each slot, its distance with all possible values is calculated, i.e., the prediction has to process 30 times, which is the number of slots in the MultiWOZ dataset. Compared with the generation methods that only process once and do not need ontology, our method is short in training efficiency and scalability. However, most task-oriented dialogue datasets contain their knowledge base containing slot value information, so it's acceptable to construct the ontology for random sampling. Besides, the results in Section \ref{T5gen} demonstrate that our method can be implemented into generation-based backbone models.}

\section*{Acknowledgements}
The work was supported in part by NSFC with Grant No. 62293482, the Basic Research Project No. HZQB-KCZYZ-2021067 of Hetao Shenzhen-HK S\&T Cooperation Zone, the National Key R\&D Program of China with grant No. 2018YFB1800800, the Shenzhen Outstanding Talents Training Fund 202002, the Guangdong Research Projects No. 2017ZT07X152, No. 2019CX01X104, and No. 2021A1515011825, the Guangdong Provincial Key Laboratory of Future Networks of Intelligence (Grant No. 2022B1212010001),  the Shenzhen Key Laboratory of Big Data and Artificial Intelligence (Grant No. ZDSYS201707251409055), and the National Key R\&D Program of China under Grant No. 2020AAA0108600.

        \bibliography{anthology_new}
	\bibliographystyle{acl_natbib}
		
	
	
		
	\end{document}